\documentclass[letterpaper, 10 pt, conference]{ieeeconf}

\usepackage{graphicx}
\usepackage{amsfonts}

\usepackage[english]{babel}

\usepackage{amsmath, amssymb}
\usepackage{float}
\usepackage{makecell}
\usepackage{multirow}
\usepackage{cite}

\usepackage{amsmath,amssymb,amsfonts}
\usepackage{algorithmic}
\usepackage{graphicx}
\usepackage{textcomp}
\usepackage{xcolor}

\usepackage{pifont}
\usepackage{subfigure}
\usepackage{epstopdf}

\usepackage{graphicx}
\usepackage{comment}
\usepackage{amsmath,amssymb} 
\usepackage{color}
\usepackage{enumerate}

\usepackage{booktabs}

\IEEEoverridecommandlockouts            

\title{\LARGE \bf
FFPA-Net: Efficient Feature Fusion with Projection Awareness for 3D Object Detection
}

\author{Chaokang Jiang, Guangming Wang, Jinxing Wu, Yanzi  Miao, and Hesheng Wang 
\thanks{*This work was supported in part by the Natural Science Foundation of China under Grant 62073222, Grant U21A20480, and Grant U1913204; in part by the Science and Technology Commission of Shanghai Municipality under Grant 21511101900; and in part by the Open Research Projects of Zhejiang Laboratory under Grant 2022NB0AB01. The first three authors contributed equally. Corresponding Author: Yanzi Miao, Hesheng Wang.}
\thanks{C. Jiang and Y. Miao is with Engineering Research Center of Intelligent Control for Underground Space, Ministry of Education, School of Information and Control Engineering, Advanced Robotics Research Center, China University of Mining and Technology, Xuzhou 221116, China.}%
\thanks{G. Wang and H. Wang are with Department of Automation, Key Laboratory of System Control and Information Processing of Ministry of Education, Key Laboratory of Marine Intelligent Equipment and System of Ministry of Education, Shanghai Engineering Research Center of Intelligent Control and Management, Shanghai Jiao Tong University, Shanghai 200240, China.}
\thanks{J. Wu is with the Department of Engineering Mechanics, Shanghai Jiao Tong University, Shanghai 200240, China.}%
}

\begin{document}

\maketitle
\thispagestyle{empty}
\pagestyle{empty}

\begin{abstract}

Promising complementarity exists between the texture features of color images and the geometric information of LiDAR point clouds. However, there still present many challenges for efficient and robust feature fusion in the field of 3D object detection. In this paper, first, unstructured 3D point clouds are filled in the 2D plane and 3D point cloud features are extracted faster using projection-aware convolution layers. Further, the corresponding indexes between different sensor signals are established in advance in the data preprocessing, which enables faster cross-modal feature fusion. To address LiDAR points and image pixels misalignment problems, two new plug-and-play fusion modules, LiCamFuse and BiLiCamFuse, are proposed. In LiCamFuse, soft query weights with perceiving the Euclidean distance of bimodal features are proposed. In BiLiCamFuse, the fusion module with dual attention is proposed to deeply correlate the geometric and textural features of the scene. The quantitative results on the KITTI dataset demonstrate that the proposed method achieves better feature-level fusion. In addition, the proposed network shows a shorter running time compared to existing methods.

\end{abstract}

\section{Introduction}
3D object detection technology is widely used in promising fields such as autonomous driving \cite{prakash2021multi,ye2022rope3d,wang2019pseudo}, SLAM \cite{zhong2018detect}, and robotics perception \cite{chen2019fast,kaur2021new}. Nowadays, accurate and efficient 3D object detection is still one of the most challenging problems in computer vision. Some works detect 3D objects utilizing various data, like monocular images\cite{ma2020rethinking,wang2021fcos3d}, stereo images\cite{wang2021plumenet, liu2021yolostereo3d}, and LiDAR point clouds\cite{shi2019pointrcnn,chen2019fast, yan2018second,vora2020pointpainting}. However, individual sensors have drawbacks. The sparsity of LiDAR point clouds and the lack of color information make it difficult to ensure high accuracy. So, fusion of multimodal features is important for achieving robust 3D detection.
\begin{figure}[t]
	\centering
	\includegraphics[scale=0.44]{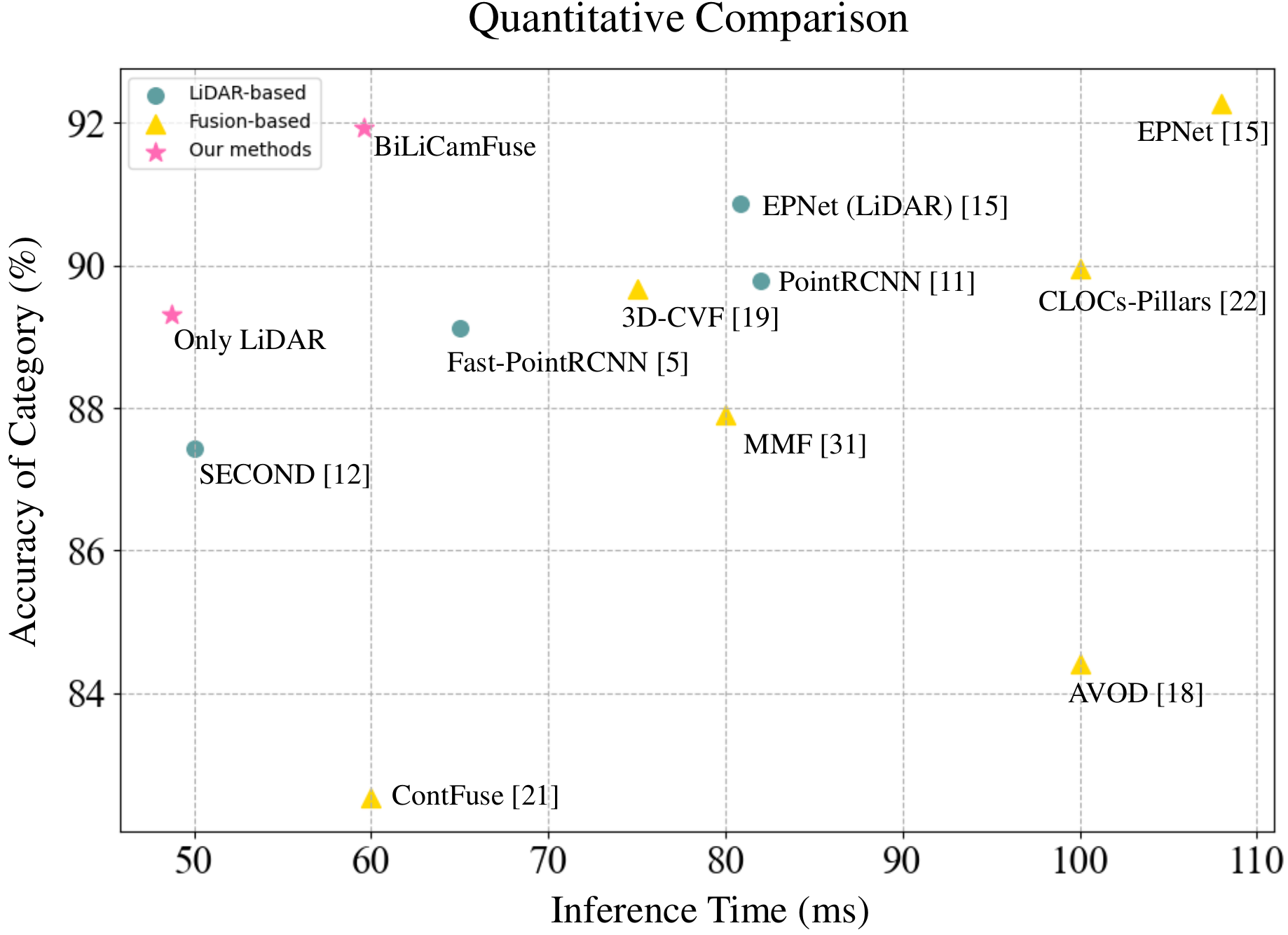}
	\vspace{-4mm}
	\caption{\textbf{Comparison of time consumption and detection precision with advanced 3D detection algorithms on the KITTI $val$ split \cite{geiger2013vision}.} The detection category is car. The 3D average precision with Intersection-Over-Union (IoU) threshold of 0.7 is used as an evaluation metric for easy scene.}
	\label{fig:infertime}
\end{figure}

Previous front view projections and BEV projections usually lose depth or height information. In the other hand, for learning point-wise features, the raw LiDAR point clouds is often sampled to 16384 points due to computational constraints \cite{shi2019pointrcnn,huang2020epnet,chen2019fast}. These methods lead to different degrees of losing the 3D geometric information of the sensor signal. In addition, it is extremely time-consuming for the farthest point sampling and $K$-nearest neighbor (KNN) query operations in deep learning methods \cite{qi2017pointnet,qi2017pointnet++,shi2019pointrcnn} that extract geometric features from the raw point cloud, especially the time consumption increases exponentially as the number of points increases. To ensure high inference speed and completeness of raw coordinates information, a novel projection method is introduced in our work. Specifically, the raw 3D coordinates information is recorded in the corresponding position in the 2D plane. This approach enables a faster feature computation with almost no loss of geometric information, which ensures that our architecture is efficient and effective. The quantitative comparisons of our approach with other methods are shown in Fig. \ref{fig:infertime}. The efficient point cloud feature extraction network and the rational 2D-3D fusion module designed in this paper make the network achieves excellent 3D object detection performance.

Several works\cite{ku2018joint,huang2020epnet,yoo20203d,qi2018frustum,liang2018deep,pang2020clocs,xie2020pi} make efforts to tackle single-modality problem by combining data from multiple sensors. The direct data fusion without information preprocessing\cite{chen2017multi,ku2018joint,qi2018frustum} neglects the different perspectives of RGB-images and point clouds. Although Liang et al.\cite{liang2018deep} use continuous convolution to solve this problem, their fusion from bird's eye view (BEV) maps is insufficient in accuracy. Another method of unidirectional fusion, EPNet \cite{huang2020epnet}, cannot deeply fuse the two complementary modalities. To alleviate cross-modal misalignment, we design two plug-and-play fusion modules, LiCamFuse and Bidirectional LiCamFuse (BiLiCamFuse). Euclidean information between the projected LiDAR coordinates and pixel coordinates is introduced into the calculation of the soft query weights in the LiCamFuse module, which improves the misalignment between the different modalities. Inspired by cost volume presented in PWC-Net\cite{sun2018pwc}, We design BiLiCamFuse, a bi-directional fusion method for augmenting the semantic information of point clouds and the geometric information of image pixels. The LiDAR points query not only the geometric features of their neighboring points but also the texture features of their neighboring pixels through the designed local KNN. We designed BiLiCamFuse, a bi-directional fusion method for augmenting the semantic information of point clouds and the geometric information of image pixels.

Our main contributions are as follows:

\begin{itemize}
\item An efficient and accurate 3D object detection architecture, FFPA-Net, is proposed. The introduced query of local KNN in the projection map significantly improves the efficiency of geometric feature extraction.

\item A new data pre-processing method is proposed to achieve more efficient fusion of the different signal features. The indexes between different sensor signals are established in advance and stored in a map, while synchronized sampling provides fast and accurate query correspondence for feature fusion.

\item In our network, multiple methods are explored to achieve cross-modal feature fusion more reasonably and efficiently, including soft query weights with perceiving the Euclidean distance of bimodal features, and fusion modules based on dual attention correlating the geometric features and texture features of the scene.

\item Our proposed 3D object detection architecture is carefully evaluated over the popular KITTI benchmark dataset. The quantitative results show the effectiveness of our strategy in this paper. Further, the proposed 2D-3D fusion network achieves an inference speed of 17 FPS on a single NVIDIA RTX 2080Ti GPU, which is 46\% less time-consuming than the EPNet \cite{huang2020epnet}, a high-performance 3D object detection network.

\end{itemize}

\section{Related Work}
\subsection{Deep Learning for Point Clouds}
As a pioneer, PointNet\cite{qi2017pointnet} proposes a simple but efficient method to obtain 3D geometric data from point clouds directly. PointNet++\cite{qi2017pointnet++} offers a hierarchical network by learning local features utilizing 3D Euclidean information and uses PointNet \cite{qi2017pointnet} recursively on different point sets. Recently, some methods in other areas are introduced into deep learning for point clouds. PCT\cite{guo2021pct} makes use of the inherent order invariance of the transformer, applying attention mechanism for data learning, and improving its accuracy on feature extraction for point clouds by using implicit Laplace operator and normalization. EfficientLO-Net\cite{wang2021efficient} uses a projection-aware representation of point clouds to organize the unordered data and applies an embedding mask on pose warping.

\subsection{3D Object Detection}

How to process the point clouds data has been a critical concern in 3D object detection. Some previous works \cite{ku2018joint, chen2017multi} focus on projection. Ku et al.\cite{ku2018joint} use BEV and image as input to extract high-resolution feature maps. The region proposal network (RPN) is used to obtain region proposals for classes, maintaining great performance with a low GPU usage. Similarly, MV3D\cite{chen2017multi} generates 3D candidate boxes from BEV and projects them to three views, taking the front view of point clouds and images as reference for the region of interest pooling (ROI-Pooling). And finally the classifier and 3D box regressor are generated from a region-based fusion network. However, the above works are still based on the 2D object detection frameworks and cause much loss of 3D information. PointNet\cite{qi2017pointnet} and PointRCNN\cite{shi2019pointrcnn} directly take raw point clouds as input, fully utilizing the 3D information. Meanwhile, voxelization \cite{shi2021pv,garcia2019geometry,mao2021voxel} is also an approach to learn distinctive features from irregular points.

To extract the characteristics of different types of data to achieve accurate and robust results on object detection, data fusion became a crucial work. PI-RCNN\cite{xie2020pi} extracts full-resolution semantic features as input in 3D detection. EPNet\cite{huang2020epnet} constructs the point-wise correspondence between the raw point clouds and image data to enhance the semantic characteristics. Prakash et al.\cite{prakash2021multi} design TransFuser to integrate LiDAR data and images using attention and add the global context into distinct modalities layer by layer. 3D-CVF \cite{yoo20203d} uses an automatic calibration projection to convert 2D camera features into smooth spatial feature maps with the highest correspondence to LiDAR features in BEV.

\section{Problem Definition}

The task for 3D object detection estimates the size, orientation, 3D position, and category of visible objects from provided sensor signals. The data input of our network are RGB image $ I= \{\{x_i^I,F_i^I\}\|i=1,\ldots,N\}$ and point clouds $PC = \{ \{x_i^L,F_i^L\}|i=1,\ldots,N\}  $, where $x^I_i \in {\mathbb{R}^2}$ is the 2D coordinates of the image pixel, and $F_i^I \in {\mathbb{R}^C}$ is the image features; $x^L_i \in {\mathbb{R}^3}$ is the 3D coordinates of the LiDAR point, and $F^L_i$ is the LiDAR features. Each bounding box of the network output is represented by a vector $\vec{v} = \{ x,y,z,h,l,w,\phi\}$, in which $x,y,z$ represent the coordinates of the center point of the box; $h,l,w$ is the height, length, and width of the box respectively; $\phi$ is the heading angle of the box.

\begin{figure*}[t]
	\centering
	\includegraphics[scale=0.575]{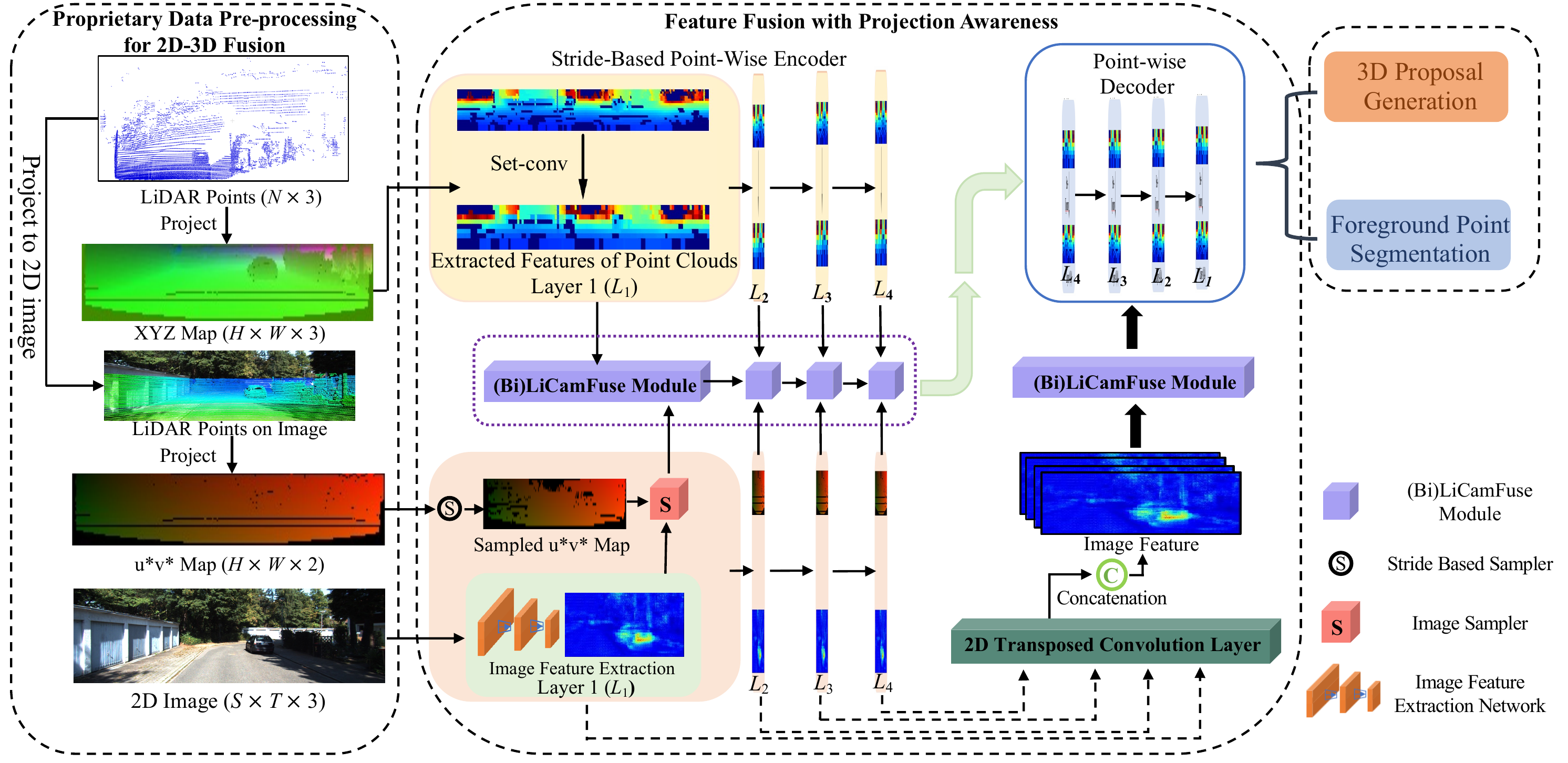}
	\vspace{-3mm}
	\caption{\textbf{The framework of our proposed FFPA-Net. }
	Firstly the raw information is organized in uniform data format by utilizing projection. Stride-based point-wise encoder serves as feature extraction to generate aligned feature representation, which is fed to (Bi)LiCamFuse module for data fusion. Finally, 3D proposals generation and foreground point segmentation are operated in the end of the detection task.}
	\label{fig:Network}
\end{figure*}

\section{Efficient Detector With 2D-3D Fusion} 

The overview of the model architecture is illustrated in Fig. \ref{fig:Network}. In FFPA-Net, the input sensor signal is first pre-processed. Then, geometric features are extracted by the stride-based point cloud feature encoder. Texture features are extracted by the convolutional network. Finally, all the feature representations are fed to the (Bi)LiCamFuse module for cross-modal feature fusion. With the fusion features, 3D proposal generation and foreground point segmentation are processed to ultimately solve the detection task.

\subsection{Proprietary Data Pre-processing for 2D-3D Fusion}
The 3D coordinates of the raw LiDAR point are stored orderly on the 2D plane. The formulas\cite{wang2021efficient} is as follows:
\begin{equation}
\vspace{-3mm}
    u = arctan(2y/x)/ \Delta \theta,
    \label{equ:one}
\end{equation}

\begin{equation}
    v = arcsin(z/\sqrt{x^2+y^2+z^2})/ \Delta \phi,
     \label{equ:two}
\end{equation}
where $x,y,z$ are the 3D coordinates of the LiDAR points. $\Delta \theta, \Delta \phi$ represent the horizontal and vertical resolution respectively. $u$ and $v$ represent the projection coordinates of the points in the LiDAR point clouds on the 2D plane. This 2D plane is called the XYZ map. Such an operation organizes the unordered 3D point data in an ordered format. The dimensions of the XYZ map are $H \times W \times 3$, as shown in Fig. \ref{fig:Network}. Next, the corresponding relationships between LiDAR points and image pixels are obtained by projecting LiDAR into the image coordinate system through a mapping matrix, which is derived from the camera intrinsics and extrinsics. The projected coordinates onto the color image from the LiDAR point are represented by ($u^*$, $v^*$). However, the stride-based downsampling operation in layer-by-layer point cloud feature learning breaks the one-to-one correspondence between the coordinates $XYZ$ of the raw point cloud and the projected coordinates ($u^*$, $v^*$) onto the color image from the point cloud.

To ensure LiDAR points and pixel points always correspond during feature fusion, we fill in the position of the map copied from the XYZ map with ($u^*$, $v^*$). This generates a $u^*v^*$ map with dimension $H \times W \times 2$, as shown in Figure \ref{fig:Network}. Such a method of creating different maps and maintaining synchronized indexes ensures that LiDAR points and pixels always maintain their original correspondence without re-creating indexes at each layer, where the different maps are $u^*,v^*$ map and XYZ map, respectively. Therefore, this method is accurate and takes less running time.

\subsection{Stride-Based Point-Wise Encoder}
In our network, sliding kernels \cite{wang2021efficient} are set to sample center points and extract features by grouping neighbor points utilizing KNN in the kernels, which is more efficient. 

\subsubsection{Stride-Based Sampling} 

In our sampling scheme, stride-based kernels are set on the XYZ map, and the kernel center points are selected, which generate the indexes of sampling points directly. This method achieves uniform distribution of sampling center points and faster speed.

\subsubsection{Feature Extraction}

In kernel, the points beyond a fixed range from the center point are discarded, and those within it are reserved. The grouping points are obtained for each center point.  With the center points and the grouping points, the features can be calculated by the following formula:
\begin{equation}
    f_{ip} = \mathop{MAXPOOL}\limits_{k=1,2,\dots,K}(FC(x_{i}^k-x_i)\oplus f_{ip}^k \oplus f_{ip}^s),
\end{equation}
in which $x_i$ represents the index of $i$-th center point, $x_i^k$ is the index of $k$-th grouping point around the $i$-th center point, and $K$ is the total number of the grouping points for each center point. $f_i^s$ and $f_i^k$ represent the feature of $i$-th center point and $k$-th grouping point around the $i$-th center point respectively. $f_i$ is the extracted feature of $i$-th center point. $\oplus$ indicates the concatenation of two tensors. $FC(\cdot)$ denotes a fully connected layer, and $MAXPOOL(\cdot)$ means the max-pooling layer.

Likewise, the stride-based sampling is applied to the pre-generated $u^*v^*$ map, which ensures that the 3D LiDAR points saved in the XYZ map and the index ($u^*$, $v^*$) of each LiDAR point in the image space saved in the $u^*v^*$ map always correspond. This provides a correspondence basis for the cross-modal feature fusion later. This process is indicated in Fig. \ref{fig:Network}. The stride-based sampling is both processed four times on the XYZ map and $u^*v^*$ map, respectively, which generates multi-resolution feature representations. 

\subsection{LiDAR Image Fusion Module}

\begin{figure}[t]
	\centering
	\includegraphics[scale=0.375]{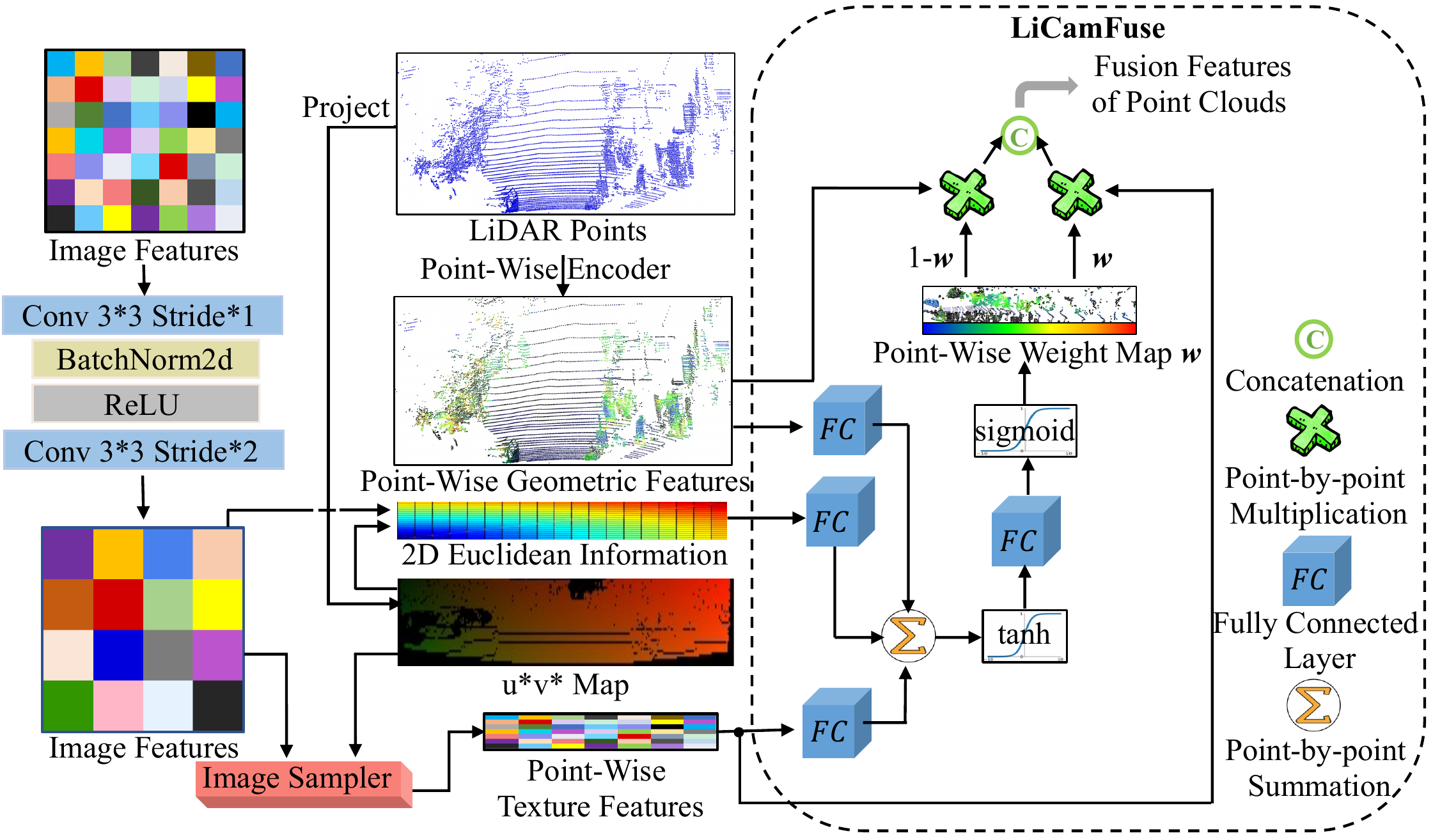}
	\vspace{-3mm}
	\caption{\textbf{The detailed network of our proposed LiCamFuse.} 
	The geometric features, texture features, and 2D Euclidean information are fused to generate matching weights for the two signals, which achieves adaptive feature fusion.}
	\label{fig:Fusion}
\end{figure}

We introduce Euclidean information into calculating point-wise weight maps to better leverage the correspondence between LiDAR features and image features. Two feature fusion modules are proposed in our work: LiCamFuse module offers an approach to fuse features moderately with high calculating speed; BiLiCamFuse module deeply fuses these two modalities in a bidirectional manner but with a little more calculation time. Both modules are plug-and-play, serving as feasible options selected according to the requirement of the detection task.

\subsubsection{Image Feature Extraction}
Image feature $F^I$ extraction is produced by 2D image convolution. The learning process is described as shown below:
\begin{scriptsize}
\begin{equation}
    F^I = Conv_{3\times3}^{2}\bigg(ReLU\Big(BN\big(Conv_{3\times3}^{1}(I)\big)\Big)\bigg), F^I \in {\mathbb{R}^{\frac{S}{4} \times \frac{T}{4} \times C}}.
\end{equation}
\end{scriptsize}
$Conv_{3\times3}^{2}$ and $Conv_{3\times3}^{1}$ are both 2D convolution. The superscript denotes the stride, and the subscript denotes the kernel size. $C$ is the number of features' dimensions. The architecture is shown in Fig. \ref{fig:Fusion}. 

\subsubsection{LiCamFuse Module}
To obtain texture features corresponding to LiDAR points in a point-wise manner, the image features after extraction are required to be indexed utilizing bi-linear interpolation. The sampled $u^*v^*$ map, which offers the corresponding relationships between LiDAR points and image, severing as an indexing map for the image sampler. After sampling, the point-wise texture features of $N \times C$ are produced, where $N=H \times W$, as shown in Fig. \ref{fig:Fusion}

The 2D coordinates obtained by projecting LiDAR points into the 2D image space ideally match the pixel coordinates of the image. However, they are often misaligned in the real world. Being able to perceive this misalignment is necessary for the process of fusing different modal features. Therefore, the proposed Euclidean information containing the differences between coordinates of points in different signals is introduced into the feature fusion module. For example, the 2D coordinates of point is $x_i=(u,v)$, and the corresponding pixel coordinates is $y_i=(u^{'}, v^{'})$. The 2D Euclidean information is generated by the followings:
\begin{equation}
    F^E_i = x_i \oplus y_i \oplus (x_i-y_i) \oplus \left \| {x_i-y_i} \right \| ,
\end{equation}
where $\left \| {\cdot} \right \|$ means the Euclidean distance between two points. Therefore, the 2D Euclidean information is a tensor of $N \times 7$.

For each point, with the point-wise geometric features $F^L_i$, the point-wise texture features $F^I_i$, and the 2D Euclidean information $F^E_i$, the feature fusion can be performed by the following formulas:

\begin{equation}
    F^F_i = FC(F^L_i)+FC(F^I_i)+FC(F^E_i),
\end{equation}
    \vspace{-4mm}
\begin{equation}
    w = sigmoid\Big(FC\big(tanh(F^F_i)\big)\Big),
\end{equation}
\begin{equation}
    F_i = w \odot F^I_i +(1 \ominus w) \odot F^P_i,
\end{equation}
where $\odot$ indicates dot product, and $\ominus$ means element-wise minus. $F_i$ denotes the fusion feature of the point.

\subsubsection{BiLiCamFuse Module} \label{BiLiCamFuse}
To fuse LiDAR and image more rationally at a deeper level, for each selected point, the BiLiCamFuse module queries not only the neighboring LiDAR points but also the neighboring image pixels. In addition, 2D Euclidean information is introduced for correcting coordinate errors between LiDAR points and image pixel points, which ultimately generates point-wise features from the information of LiDAR points and image points. 

The input of one BiLiCamFuse module includes two sets of point coordinates $x^L, x^I$ and point features $F^L, F^I$. $x^L \in {\mathbb{R}^3}$ represents the 3D coordinates of the LiDAR point; $x^I \in {\mathbb{R}^2}$ denotes the 2D coordinates of the image pixel; $F^L, F^I$ are the corresponding features respectively. The module outputs ${\mathcal{O}}=\{o_i=\{x_i, F^O_i\}|i=1,\cdots,N\}$, where $F^O_i \in {\mathbb{R}^{C'}}$ is the point-wise fusion feature.

\begin{figure}[t]
	\centering
	\includegraphics[scale=0.48]{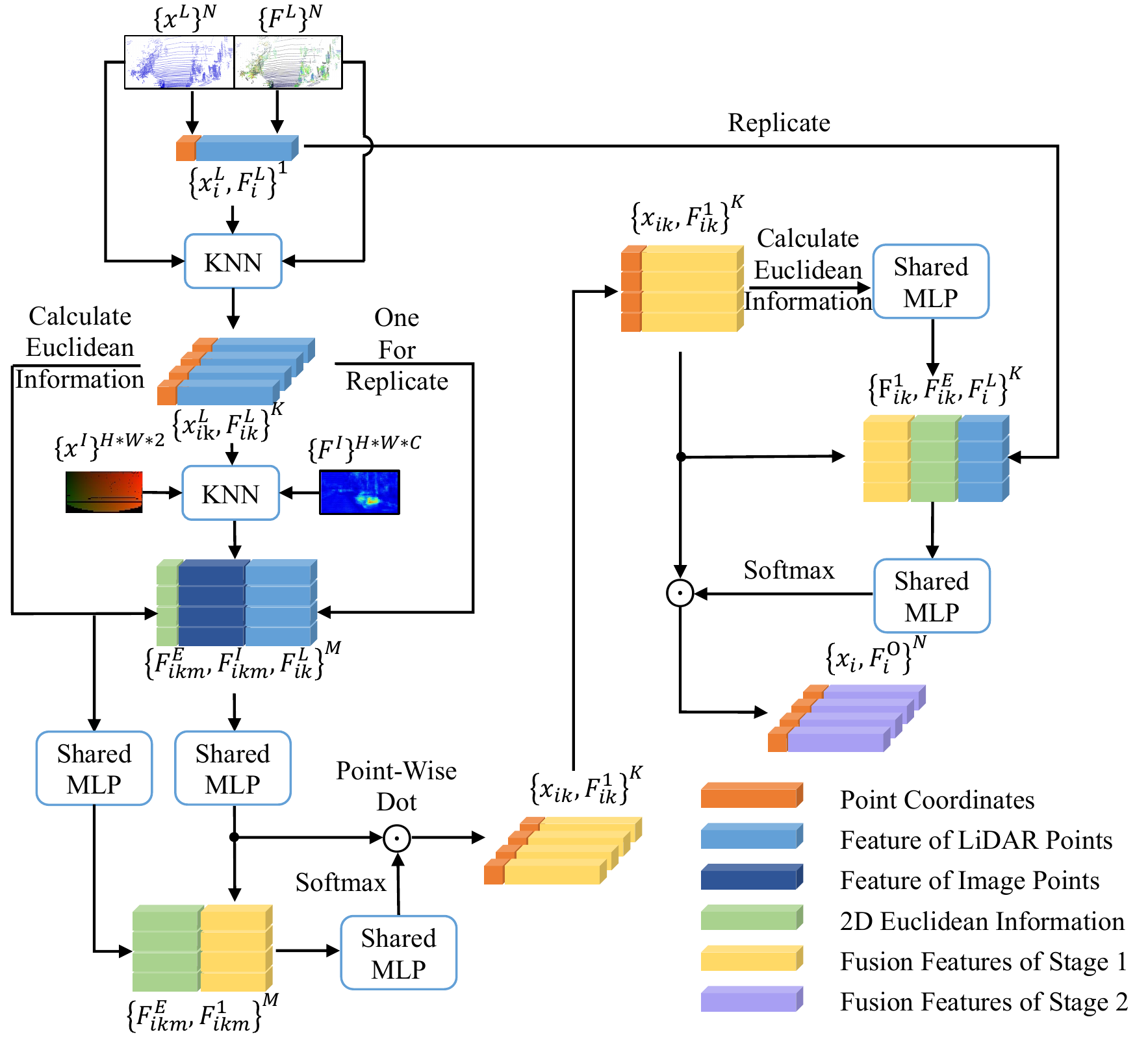}
	\vspace{-6mm}
	\caption{\textbf{The semi-architecture of our BiLiCamFuse.} 
	In this network, LiDAR points search image points for feature augmenting. Neighbor points are queried for LiDAR points both in the point clouds and image pixels. And the feature fusion incorporating Euclidean information is operated twice based on the self-adaptive attentive weights. The same fusion operation is applied on the image points in the other semi-module.}
	\label{fig:Fusion2}
\end{figure}

Fig. \ref{fig:Fusion2} indicates the detailed fusion process. Each LiDAR point first queries $K$ nearest neighbors with the representation $p_{ik}=\{x_{ik}^L, F_{ik}^L\}$ in XYZ map, where $k$-th point is one of the $K$ neighboring points in XYZ map. And each point $p_{ik}$ queries $M$ nearest neighbors in $u^* v^*$ map with the indexed image feature $F_{ikm}^I$, in which $m$-th point is one of the $M$ neighboring points in $u^* v^*$ map. Actually, before the second KNN is performed, the correspondence between XYZ map and $u^* v^*$ map is utilized to shift 3D LiDAR points to the image pixels. Moreover, 2D Euclidean information $F^E_{ikm}$ is calculated between $p_{ik}$ and its $M$ nearest neighbors in $u^* v^*$ map, and the feature $F_{ik}^L$ is duplicated $M$ times. The features above are concatenated to generate indexed feature set $\{F^E_{ikm},F^I_{ikm},F^L_{ik}\}$.

The fusion features of stage 1 for each selected point in $u^*v^*$ map are calculated by the following formula:
\begin{equation}
    F^1_{ikm} = MLP(F^E_{ikm}\oplus F^I_{ikm}\oplus F^L_{ik}),
\end{equation}
where MLP denotes Muti-Layer Perceptron, and $F^1_{ikm} \in {\mathbb{R}^{c'}}$. The 2D Euclidean information is also used in the soft query weights learning. Therefore, for each point selected in LiDAR points, the fusion feature is:
\begin{small}
\begin{equation}
    F^1_{ik} = F^1_{ikm} \odot softmax\Big(MLP\big(MLP(F^E_{ikm}\oplus F^1_{ikm})\big)\Big),
\end{equation}
\end{small}
where softmax is used to normalize the weight.

The same procedure as the stage 1 needs to be conducted again to learn the feature representation of each LiDAR point. The fusion features of stage 1 are combined, and the feature of the LiDAR point  $F_i^L$ is duplicated $K$ times. Meanwhile, 2D Euclidean information $F^E_{ik}$ is calculated between $p_i$ and its $K$ nearest neighbors in LiDAR points. Therefore, the fusion features of stage 2, also the output feature ${F_i^O}$ for each LiDAR point is:
\begin{equation}
    F^O_i = F^1_{ik} \odot softmax\big(MLP(F^1_{ik}\oplus F^E_{ik}\oplus F^L_{i})\big).
\end{equation}

The interactive query mode provides a more accurate feature representation for LiDAR points. To achieve a more reasonable deep feature fusion, the same operation performed on the LiDAR points is employed on the image points, which means querying $K$ neighbors in the image pixels for each image point and $M$ neighbors in the LiDAR points for selected image neighbors, and fusion is processed. Therefore, the fusion feature for each image pixel $F^{O'}_i$ is obtained. The final feature for each point is:
\begin{equation}
    F_i = ReLU\Big(BN\big(Conv^1_{1\times 1}(F^O_i\oplus F^{O'}_i)\big)\Big).
\end{equation}
After the feature encoding and feature fusion of the point clouds is completed, the LiDAR points are restored to the resolution of the original size using an interpolation-based decoder. Finally, we further fuse the image features at the level of the point cloud resolution on the original size.

\subsection{Loss Function}
In this paper, the existing supervised loss \cite{huang2020epnet, shi2019pointrcnn} function is adopted to supervise the proposed 3D object detection network. The total losses are as follows:
\begin{equation}
    \mathcal{L}_{total} = \mathcal{L}_{rpn}+\mathcal{L}_{rcnn},
\end{equation}
\begin{equation}
    \mathcal{L}_{rpn} = \mathcal{L}_{cls} + \mathcal{L}_{reg} + \lambda \mathcal{L}_{cf},
\end{equation}

where $\mathcal{L}_{rpn}$ and $\mathcal{L}_{rcnn}$ represent the loss of RPN and the loss of the refinement network. Both losses include classification loss, regression loss, and consistency enforcing loss.

\section{Experients}

\setlength{\tabcolsep}{3.1mm}
\begin{table*}[t]
	\begin{center}
		\caption{\textbf{Quantitative results of 3D object detection on the KITTI test benchmark.} We adopt the latest 40 sampling recall points to evaluate the 2D, 3D, and BEV accuracy for the car category, where the 3D IoU threshold for recall is set to 0.7. The LiDAR signal and color image of the input to the network are represented by ``L'' and ``I'', respectively. The inference time of each method on the 2080Ti GPU device is also given in the table. ``$\sim$'' means the runtime on the 1080Ti GPU.}
		\label{table:one}
		\begin{tabular}{cccccccccccc}
\toprule
&  &  &\multicolumn{3}{c}{\centering{2D (IoU Threshold: 0.7)}}&\multicolumn{3}{c}{\centering{3D (IoU Threshold: 0.7)}}&\multicolumn{3}{c}{\centering{BEV (IoU Threshold: 0.7)}}\\\cmidrule(lr){4-6}\cmidrule(lr){7-9}\cmidrule(lr){10-12}
\multirow{-2}{*}{Method}         & \multirow{-2}{*}{Sensor(s)}          & \multirow{-2}{*}{Time (ms)}  & Easy                         & Mod                          & Hard                         & Easy                         & Mod                          & Hard                         & Easy                         & Mod                          & Hard \\ \midrule
PatchNet \cite{ma2020rethinking} & I & $\sim$400 & -- & -- & -- &15.68 & 11.12 & 10.17 & 22.97 & 16.86 & 14.97 \\
MV3D \cite{chen2017multi} & L & 240 & 89.80 & 79.76 & 78.61 &74.97 & 63.63 & 54.00 & 86.62 & 78.93 & 69.80 \\
PointRCNN \cite{shi2019pointrcnn} & L & 100 & -- & -- & -- &86.96 &  75.64 &  70.7 &  92.13 &  87.39 &  82.72 \\
SECOND \cite{yan2018second} & L & 50  &  90.40  &  88.40 & 80.21 &  83.13 &  73.66  & 66.20  & 88.07  & 79.37 &  77.95 \\
Fast PointRCNN \cite{chen2019fast} & L & 65 & -- & -- & --  & 85.29 &  77.40 &  70.24 & 90.87  & 87.84  & 80.52 \\
PointPainting \cite{vora2020pointpainting} & L & -- & -- & -- & --  & 82.58 & 74.31 & 68.99 & 86.10 & 88.35 & 79.83 \\
\midrule
MV3D \cite{chen2017multi} & L+I & 360 & 90.53& 89.17& 80.16 &71.09 &62.35 &55.12& 86.02& 76.90 &68.49 \\
F-PointNet \cite{qi2018frustum} & L+I & 170  & 90.78 & 90.00 & 80.80 & 82.19                        & 69.79                        & 60.59                        & 91.17                        & 84.67                        & 74.77  \\
AVOD \cite{ku2018joint} & L+I & 100  & 89.73 &  88.08 &  80.14 & 83.07 & 71.76 & 65.73 & 89.75 & 84.95 & 78.32 \\
ContFuse \cite{liang2018deep} & L+I & 60  & -- & -- & -- & 82.54 & 66.22 & 64.04 & 88.81 & 85.83 & 77.3 \\

MMF \cite{liang2019multi} & L+I & 80 & 91.82 & 90.17 & \bf88.54 & 86.81 & 76.75 & 68.41 & 89.49 & 87.47 & 79.10 \\
CLOCs-SecCas \cite{pang2020clocs} & L+I & 100 & -- & -- & -- & 86.38 & \bf78.45 & \bf72.45 & 91.16 & \bf88.23 & \bf82.63    \\
PI-RCNN \cite{xie2020pi} & L+I & -- & -- & -- & -- & 84.37 & 74.82 & 70.03 & -- &-- & -- \\
FFPA-Net (Ours) & L+I  & \bf59  & \bf96.08 & \bf90.31 & 82.91 & \textbf{87.14}  & 72.89 & 65.19 & \bf92.12 & 85.63  & 76.15        \\ \bottomrule
		\end{tabular}
	\end{center}
\end{table*}

\subsection{Setup}

\subsubsection{Implementation Details}
The depth, width, and height of the point clouds fed into the network are limited to [0, 70.4] meters, [-40, 40] meters, and [-1, 3] meters, respectively. The sizes $S$ and $T$ of the 2D images are padded to 1280 $\times$ 384. The feature extraction consists of four downsamples, where the point clouds feature dimensions of each layer are 128, 256, 512, 1024, and the image feature dimensions of each layer are 64, 128, 256, 512, respectively. The kernel sizes based on the stride sampling module are [9, 13] and [9, 5] for the first two layers and the last two layers, respectively. The horizontal and vertical sliding strides are 2. The parameter settings of the 3D box extractor are consistent with PointRCNN \cite{shi2019pointrcnn}. 

\subsection{Analysis of Experimental Results}

From Table \ref{table:one}, we compare fairly with existing methods on the KITTI test set \cite{geiger2012we}. The inference time and detection performance of each 3D object detection network are shown in Table \ref{table:one}. Compared with advanced 3D detection networks \cite{chen2017multi, vora2020pointpainting, yan2018second, chen2019fast} based on LiDAR signals such as PointRCNN \cite{shi2019pointrcnn}, our method reasonably fuses the texture information of the captured images. The proposed method achieves excellent 2D detection and 3D detection results. Notably, the inference time of our network is less compared to that of LiDAR signal-based algorithms. Less time consumption is attributed to the projection-aware feature learning of our network, which reasonably extracts local features and performs local KNN operations. The network designed in this paper achieves efficient 3D detection results. Our method detects 3D objects in a frame taking only 59 ms on an NVIDIA GeForce RTX 2080Ti GPU. As described in \ref{BiLiCamFuse}, BiLiCamFuse fuses LiDAR features and image features, more deeply. The deeper feature interaction fusion achieves high accuracy 3D object detection results.

In Table \ref{table:val}, we also provide the results of other 3D detection networks on the KITTI $val$ split \cite{geiger2012we}. Again, our method maintains excellent detection accuracy and speed on the validation dataset.  Compared with advanced bimodal fusion 3D detection networks, the proposed LiCamFuse module introduces feature Euclidean information. LiCamFuse better perceives the degree of alignment of LiDAR features and image features and assigns appropriate fusion weights to each point. The proposed method achieves excellent detection results for 3D object detection.
\vspace{-1mm}
\setlength{\tabcolsep}{1.6mm}
\begin{table}[t]
	\begin{center}
		\caption{\textbf{Comparison with advanced methods on KITTI $val$ split.}}
		\label{table:val}
		
		\begin{tabular}{cccccc}
\toprule
\multicolumn{1}{c}{\multirow{2}{*}{\begin{tabular}[c]{@{}l@{}}Method\end{tabular}}} & \multicolumn{1}{c}{\multirow{2}{*}{Sensor(s)}} & \multicolumn{3}{c}{\centering{3D (IoU Threshold: 0.7)}} & \multicolumn{1}{c}{\multirow{2}{*}{Time (ms)}} \\ \cmidrule{3-5}
\multicolumn{1}{c}{}                                                                               & & Easy         & Mod          & Hard         & \multicolumn{1}{c}{}                      \\ \midrule
PointRCNN \cite{fan2019pointrnn}& L
&88.88 & 78.63 & 77.38 & 100.00 \\
EPNet \cite{huang2020epnet} & L
& 90.87 & 81.15 & 79.59 & 80.87 \\
3D-CVF \cite{yoo20203d} & L+I
& 89.67 & 79.88 & 78.47 & $\sim$75 \\
FocalsConv-F \cite{Chen_2022_CVPR} & L+I & 89.82 & \bf85.22 & \bf85.19 & --  \\
MSF-MC \cite{wang2021multi} & L+I
& 88.14 & 77.48 & 75.92 & -- \\
Pointformer  \cite{pan20213d} & L+I
& 90.05 & 79.65 & 78.89 & -- \\
PointSIF\cite{zhao20193d} & L+I & 85.62 & 72.05 & 64.19 & -- \\
LiCamFuse (Ours) & L+I & 91.59 & 76.57 & 72.24 & \bf56.58 \\
FFPA-Net (Ours) & L+I & \bf91.93 &  77.20 & 70.44 & 59.40\\
\bottomrule
		\end{tabular}
	\end{center}
\end{table}

\subsection{Ablation Study}
\subsubsection{Ablation Experiments of The Proposed Different Fusion Components}
As shown in Table \ref{table:two}, the results on the KITTI $val$ split support the rationality of our proposed fusion module. The feature fusion weights are calculated in LiCamFuse to mitigate the maladaption of the two modalities. BiLiCamFuse is a deep point-wise feature fusion that enables LiDAR points to retain their own geometric features while querying the features of their neighboring pixels. The BiLiCamFuse achieves better fusion results but increases the network runtime.
\setlength{\tabcolsep}{1.45mm}
\begin{table}[t]
	\begin{center}
		\caption{\textbf{Ablation study of the effectiveness of the proposed module.} All methods are evaluated on the KITTI $val$ split.}
		\label{table:two}
		\begin{tabular}{ccccccc}
\toprule
\multicolumn{1}{c}{\multirow{2}{*}{\begin{tabular}[c]{@{}c@{}}LI-Fusion \\ \cite{huang2020epnet} \end{tabular}}} & \multicolumn{1}{c}{\multirow{2}{*}{\begin{tabular}[c]{@{}c@{}}LiCam\\Fuse\end{tabular}}} & \multicolumn{1}{c}{\multirow{2}{*}{\begin{tabular}[c]{@{}c@{}}BiLiCam\\Fuse\end{tabular}}} & \multicolumn{3}{c}{\centering{3D (IoU Threshold: 0.7)}} & \multirow{2}{*}{Time (ms)} \\ \cmidrule(lr){4-6}
\multicolumn{1}{c}{}                                                                      & \multicolumn{1}{c}{}                                                                             & \multicolumn{1}{c}{}                                                                      & \multicolumn{1}{c}{Easy}         & \multicolumn{1}{c}{Mod}          & \multicolumn{1}{c}{Hard}         &                       \\ \midrule
                                 \ding{55} &  \ding{55} &  \ding{55} & 89.31 & 74.21 & 70.87 & \bf48.75 \\
                                 \ding{51} &  \ding{55} & \ding{55} & 90.92 & 76.52 & 71.30 & 56.11 \\
                                 \ding{55} &  \ding{51} &  \ding{55} & 91.59 & 76.57 & \bf72.24 & 56.58 \\
                                 \ding{55} &  \ding{55} & \ding{51} & \bf91.93 &  \bf77.20 & 70.44 & 59.40 \\
                                     \bottomrule
		\end{tabular}
	\end{center}
\end{table}
\subsubsection{Impact of Projection Resolution of Point Clouds on Network Performance} 
In the data preprocessing of this paper, the resolution of the point clouds projection influences the learning effect of the network. The larger the resolution of the point clouds projection, the fewer points are lost in the projection but the more zero points in the projection plane. Conversely, the smaller the resolution of the point clouds projection, the more points are lost in the projection but the fewer zeros in the projection plane. We present the most suitable projection resolution in Table \ref{table:three}.
\setlength{\tabcolsep}{3mm}
\begin{table}[t]
	\begin{center}
		\caption{\textbf{Analysis of the effect of different projection resolutions of point clouds on the results.}}
		\label{table:three}
		
		\begin{tabular}{ccccc}
\toprule
\multicolumn{1}{c}{\multirow{2}{*}{\begin{tabular}[c]{@{}l@{}}Projected\\ Resolution\end{tabular}}} & \multicolumn{3}{c}{\centering{3D (IoU Threshold: 0.7)}} & \multicolumn{1}{c}{\multirow{2}{*}{Time (ms)}} \\ \cmidrule{2-4}
\multicolumn{1}{c}{}                                                                                & Easy         & Mod          & Hard         & \multicolumn{1}{c}{}                      \\ \midrule
37$\times$180                                                                                               & 90.76        & 73.91        & 69.71       & \bf49.68                               \\
40$\times$275                                                                                               & \textbf{91.93}        & \textbf{77.19}        & \textbf{70.43}        & 59.40 \\
46$\times$420   & 90.61        & 76.66        & 73.96        & 70.93                                      \\
\bottomrule
		\end{tabular}
	\end{center}
\end{table}

\section{Conclusions}

In this paper, an efficient and accurate 2D-3D fused 3D object detection network is designed. The proposed novel method of data pre-processing accelerates the fusion of multimodal features. Also, the introduced projection-based geometric feature extraction method is more efficient. Further, we propose two novel 2D-3D fusion modules. LiCamFuse adaptively and efficiently fuses features from different modalities, and BiLiCamFuse deeply fuses LiDAR point features and pixel point features. The proposed FFPA-Net achieves efficient and accurate 3D detection results on the popular KITTI test set and validation set.

\bibliographystyle{IEEEtran}  
\bibliography{IEEEfull,root} 

\end{document}